\documentclass[lettersize,journal]{IEEEtran}
\usepackage{amsmath,amsfonts}
\usepackage{algorithmic}
\usepackage{algorithm}
\usepackage{array}
\usepackage[caption=false,font=normalsize,labelfont=sf,textfont=sf]{subfig}
\usepackage{textcomp}
\usepackage{stfloats}
\usepackage{hyperref}
\usepackage{url}
\usepackage{verbatim}
\usepackage{booktabs}
\usepackage{graphicx}
\usepackage{cite}
\usepackage{color}
\usepackage{subcaption}
\begin{document}

\title{A real-time, robust and versatile visual-SLAM framework based on deep learning networks}

\author{Xiao Zhang$^{1,2}$, 
        Hongbin Dong$^1$,
        Haoxin Zhang$^{2,3}$,
        Xiaozhou Zhu$^2$,
        Shuaixin Li*$^2$,
        Baosong Deng*$^2$,
        
\thanks{ This work was funded by National Natural Science Foundation of China under Grant 42201501. 1.Harbin Engineering University, College Of Computer Science And Technology, National Engineering Laboratory for E-Government Modeling and Simulation. 2.Intelligent Game and Decision Laboratory, Chinese Academy of Military Science, Beijing. Corresponding author: Shuaixin Li, Co-corresponding author:  Baosong Deng. 3.Sun Yat-sen University}
}


\markboth{Journal of \LaTeX\ Class Files,~Vol.~14, No.~8, August~2021}%
{Shell \MakeLowercase{\textit{et al.}}: A Sample Article Using IEEEtran.cls for IEEE Journals}


\maketitle

\begin{abstract}
In this letter, we investigate the paradigm of deep learning techniques to enhance the performance of visual-based SLAM systems, particularly in challenging environments. By leveraging deep feature extraction and matching methods, we propose a robust, versatile hybrid visual SLAM framework, Rover-SLAM, aimed at improving adaptability in adverse conditions, such as dynamic lighting conditions, areas with weak textures, and significant camera jitter. Building on excellent learning-based algorithms of recent years, we designed from scratch a novel system that uses the same feature extraction and matching approaches for all SLAM tasks. Our system supports multiple modes, including monocular, stereo, monocular-inertial, and stereo-inertial configurations, offering flexibility to address diverse real-world scenarios. 
Through comprehensive experiments conducted on publicly available datasets and self-collected data, we demonstrate the superior performance of our Rover-SLAM system compared to the SOTA approaches. We also conducted an in-depth analysis of the integration of visual SLAM with deep learning methods quantitatively to provide insights for future research endeavors in this domain. The experimental results showcase the system's capability of achieving higher localization accuracy and robust tracking performance. To facilitate further research and foster community collaboration, we have made the source code of our system publicly available at \url{https://github.com/zzzzxxxx111/SLslam}.
\end{abstract}


\section{Introduction}
SLAM (Simultaneous Localization and Mapping) is a critical technology in the domains of robotics, autonomous driving, and 3D reconstruction. It concurrently determines the 6DoF (Degree of Freedom) sensor poses while building a 3D map of the traversed environment\cite{cadena2016past}. Vision and inertial sensors are the most commonly employed sensing devices for this purpose, and numerous solutions utilizing these sensors have been extensively studied and explored. 

Over the past decades, numerous brilliant visual-based SLAM solutions employing classical computer vision methods have emerged, including ORBSLAM\cite{mur2017orb}, 
and MSCKF\cite{MSCKF}, driving significant evolution in this domain. While many of the foundational issues have been addressed, recent researches have focused on enhancing the robustness and adaptability of SLAM under extreme conditions\cite{cadena2016past}. 
In challenging environments, such as those with dynamic lighting and areas with weak textures, traditional feature extraction algorithms, which primarily focus on local information in images and often overlook structural and semantic details, substantially limit the effectiveness of pose estimation. As a result, existing SLAM systems may struggle to obtain accurate and stable features in these scenarios, leading to unstable or ineffective tracking.
\begin{figure}[t]
    \centering
    \includegraphics[width=1\linewidth]{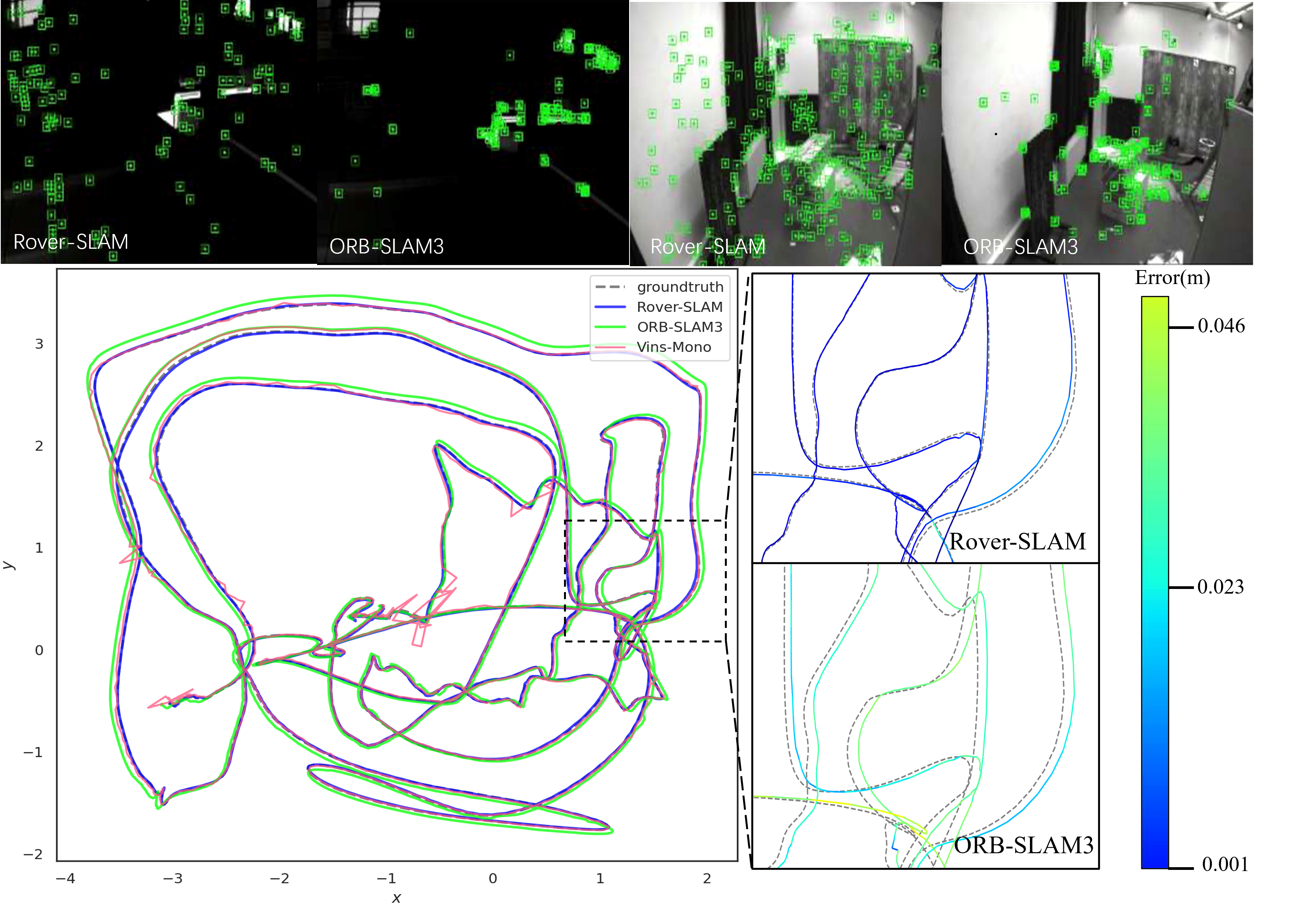}
    \caption{Rover-SLAM vs other SOTA methods. \textbf{Top}: Comparison of map point tracking performance between the proposed Rover-SLAM and ORB-SLAM3 in a challenging environment. \textbf{Bottom-Left}: Comparison of the trajectories obtained using the  ORB-SLAM3, VINS-Mono and Rover-SLAM, in a EuRoc V203 sequence,  which is the most challenging sequence for shaking and dynamic lighting reason. \textbf{Bottom-Right}: Map details color-coded with the amount of error, i.e. green corresponds to higher error levels, and blue to lower ones.}
    \label{fig:enter-label}
\end{figure}

The rapid advancement of deep learning has revolutionized the field of computer vision, enabling models to assimilate complex scene structures and semantic information through extensive data training, which has made environment perception more intelligent. The integration of deep learning with SLAM follows two routes. The first involves end-to-end algorithms based on deep learning, such as NICE-SLAM\cite{zhu2022nice}, and DVI-SLAM\cite{peng2023dvi}. These methods often incur high computational costs, long optimization times, and face challenges in real-time tracking, and the lack of detailed geometric correspondences can lead to suboptimal positioning accuracy.
The second route, known as hybrid SLAM, enhances specific SLAM modules using deep learning. This approach balances traditional geometric methods and deep learning techniques, leveraging their strengths to balance geometric constraints and semantic understanding.
Despite their advantages, existing hybrid SLAM systems have limitations. For instance, DX-Net\cite{2020dxslam} simply replaces ORB features with deep features but relies on traditional methods for tracking, leading to incoherence in the deep feature information. Similarly, SP-Loop\cite{Wang_Xu_Fan_Xiang_2023} integrates deep learning features only into the loop closure module, retaining traditional feature extraction methods elsewhere. Consequently, there is a need for a versatile hybrid SLAM method that effectively integrates deep learning technology to address complex environmental challenges comprehensively.

In this work, we propose Rover-SLAM, a real-time, robust, and versatile visual-SLAM framework. We incorporate a SOTA learning-based feature extraction module, SuperPoint\cite{detone2018superpoint}, as the sole representation form throughout the system. Additionally, 
we replace traditional feature matching methods with the learning-based approach LightGlue\cite{lightglue}, which enhances matching performance under significant changes in viewing angles. We preprocess the learning-based feature descriptors to align with the training of the corresponding visual vocabulary and design a strategic feature selection mechanism to maintain accuracy and efficiency. 
Finally, we conduct a series of experiments to demonstrate that the proposed SLAM framework improves both trajectory estimation accuracy and tracking robustness in various challenging scenarios (see Fig.\ref{fig:enter-label}). Our contributions in this work include the following: 
\begin{itemize}
\item We propose a real-time, learning-based vSLAM (visual-based SLAM) framework designed to enhance adaptability in challenging environments. By integrating deep learning modules and novel combination schemes throughout the SLAM system, covering tracking, local mapping, and loop closure, we significantly improve both the robustness and accuracy of SLAM system while maintaining real-time performance.
\item We developed the first versatile SLAM system that incorporates deep feature extraction and matching approaches. The proposed system supports various sensor configurations, including monocular, stereo, monocular-inertial, stereo-inertial, catering to diverse application needs.
\item We conducted extensive experiments to demonstrate the effectiveness and robustness of our system. Results from both public datasets and self-collected datasets indicate its superiority over other state-of-the-art SLAM systems. The system is fully implemented in C++ and ONNX, ensuring real-time operation. For the benefit of the research community, the source code is available at \url{https://github.com/zzzzxxxx111/SLslam}. 
\end{itemize}
\section{Related Work}
\subsection{Traditional visual SLAM works}
The development of SLAM has been significantly influenced by traditional geometric methodologies, resulting in numerous classic and successful approaches. In particular, DSO (direct sparse odometry)\cite{dso} exemplifies direct SLAM by sampling pixels with intensity gradients in images and integrating a photometric error model and all model parameters into a joint optimization function. ORB-SLAM\cite{mur2017orb}, a representative sparse feature-based SLAM system, utilizes ORB features to describe scenes. 
With the growing trend of assisting the vision system with a low-cost IMU (inertial measurement unit), the VINS (visual-inertial system) has gained prominence. 
VINS-Mono\cite{vins} integrates pre-integrated IMU measurements and feature observations to achieve high-precision visual-inertial odometry (VIO), forming a complete monocular VIO-SLAM system with loop closure detection and graph optimization. ORB-SLAM3\cite{orbslam3} extends the capabilities of ORB-SLAM by incorporating inertial data to improve performance.

A common issue with these traditional image processing-based SLAM systems is their susceptibility to challenging environments, which can lead to inaccurate pose estimation or even lost tracking. 
\subsection{Deep learning-based visual SLAM works}
To address the aforementioned issues, researchers are increasingly incorporating deep learning methodologies into SLAM systems to enhance robustness in challenging scenarios. One approach involves training end-to-end models for visual SLAM\cite{d3vo}. For instance, UnDeepVO\cite{undeepvo} is notable as the first end-to-end visual odometry (VO) framework based on neural networks, leveraging stereo training to recover absolute scale by utilizing spatial and temporal geometric constraints.  
DROID-SLAM\cite{teed2021droid} introduces a micro-loop optimization heuristic to enhance the network performance and generalization capabilities. While these end-to-end solutions offer adaptability across various application scenarios, they typically require extensive datasets and significant computing resources for training. In addition, they may experience performance degradation when transitioning to different scenarios.

Another approach is to integrate learning-based methods into specific SLAM systems, known as hybrid SLAM\cite{gcnv2}. Lift-SLAM\cite{2021liftslam} incorporates the learned invariant feature transform (LIFT) network for feature extraction within the ORB-SLAM system's backend. Similarly, 
DX-SLAM\cite{2020dxslam} replaces ORB features in ORB-SLAM with learning-based features and uses the OpenVINO toolkit for model deployment. However, these approaches still rely on traditional methods for feature matching during tracking. AirVO\cite{airvo} uses a learning-based feature points extraction and matching method, optimizing the pose by jointly minimizing point and line feature reprojection errors to improve estimation accuracy. Nonetheless, AirVO is not a complete SLAM system and is limited to stereo accessible environments.
\section{Method}
\subsection{The System Overview}
\begin{figure}[tb] 
  \includegraphics[width=0.5\textwidth]{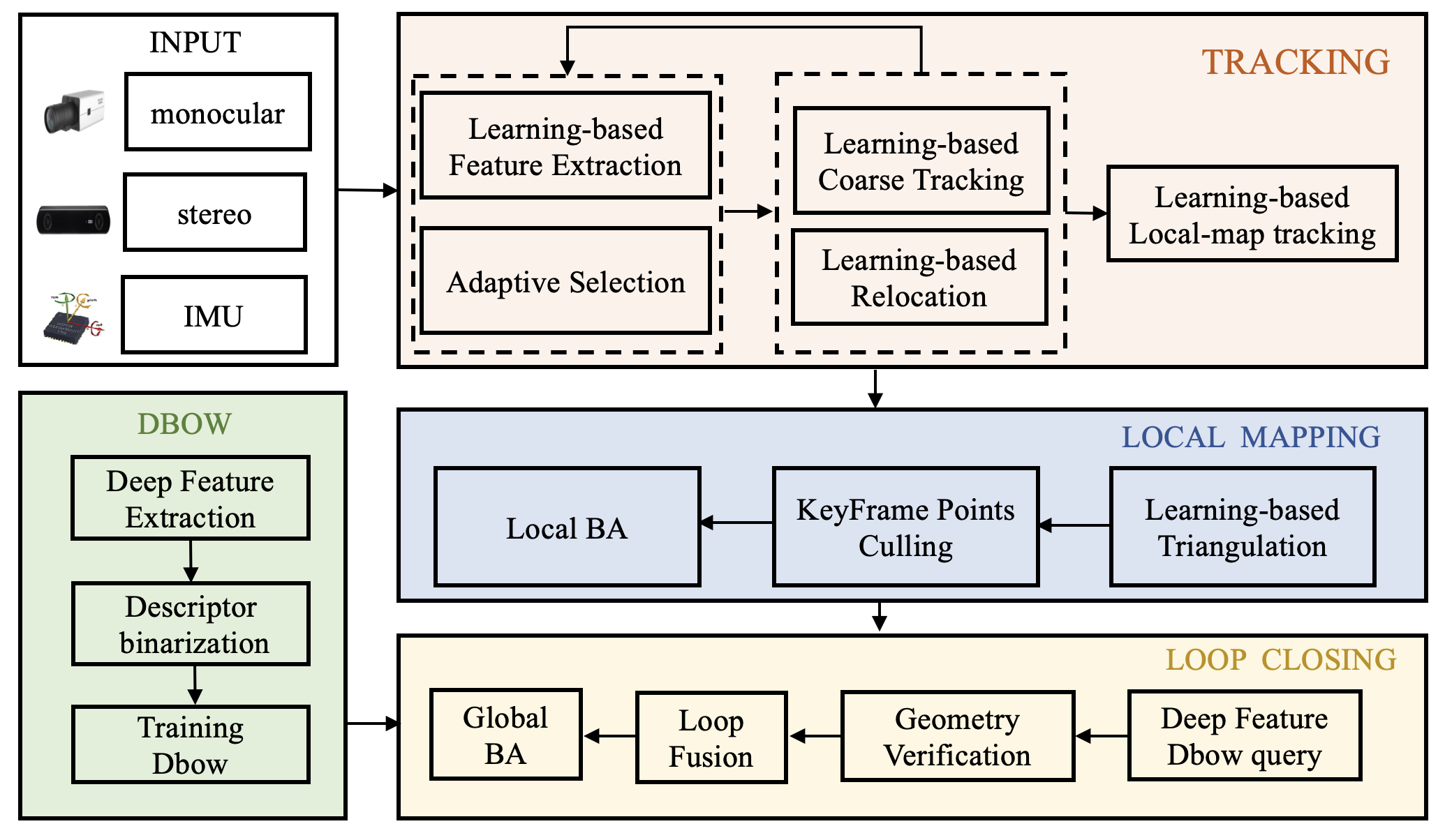}
  \caption{The framework of Rover-SLAM.The system consists of the tracking, local mapping and loop closing
modules.}
  \label{SystemOverview}
\end{figure}
The pipeline of the proposed Rover-SLAM builds upon the standard parallel vSLAM paradigm\cite{klein2009parallel}, and follows the architectural design of ORB-SLAM3 (see Fig.\ref{SystemOverview}). The system is divided into three modules, i.e. tracking, local mapping, and loop closure, and supports four configurations, i.e. monocular, monocular-inertial, stereo, and stereo-inertial. In our implementation, the ONNX Runtime deep learning deployment framework is employed, incorporating the SOTA learning-based methods, SuperPoint and LightGlue. Notably, our framework is flexible and not restricted to specific deep learning models, allowing for replacements such as XFeat\cite{potje2024xfeat} or LoFTR\cite{sun2021loftr}. 
\subsection{Adaptive Feature Extraction}
\begin{figure}[tb]
    \centering
    \includegraphics[width=1\linewidth]{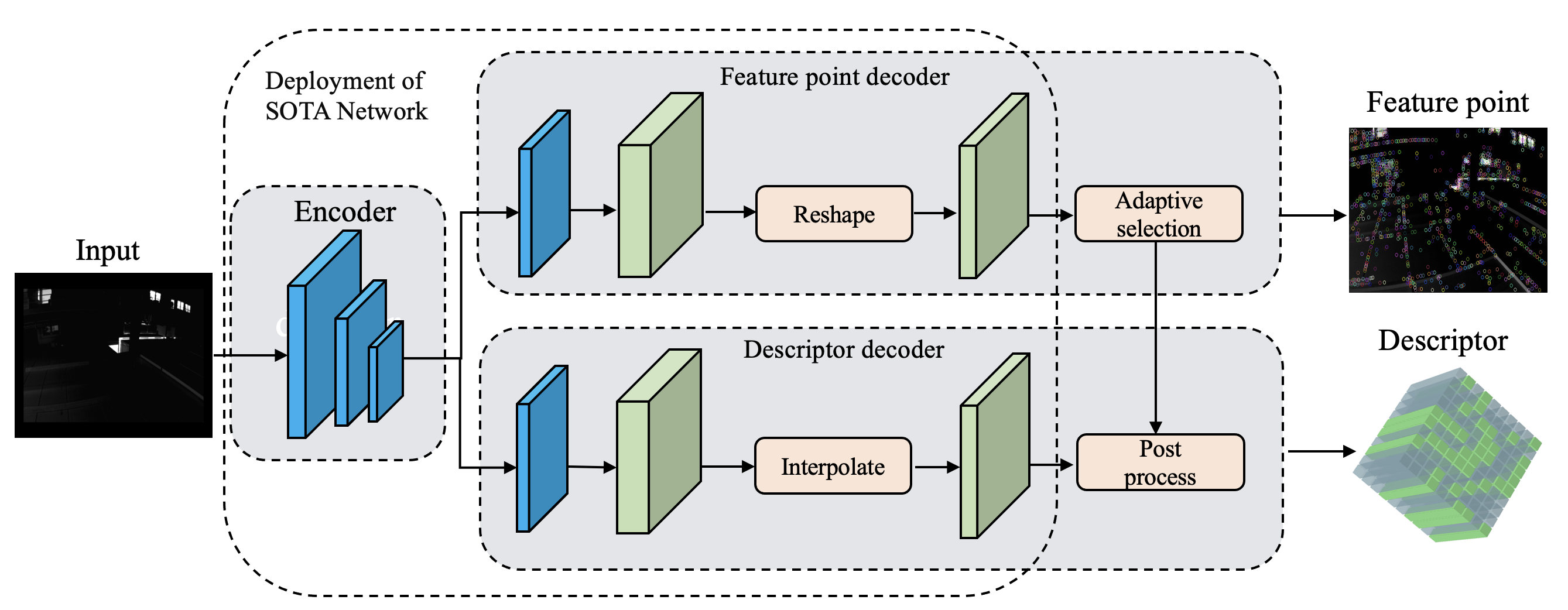}
    \caption{Overview of feature extraction. Using SuperPoint as the feature extraction network, which includes an encoder, a feature decoder, and a descriptor decoder. Subsequently, an adaptive filter is applied to sort out feature points. }
    \label{superpoint}
\end{figure}

The proposed framework employs the effective SuperPoint feature extraction network. This network comprises a shared encoder, a feature detection decoder, and a descriptor decoder, outputting a probability tensor of features and a descriptor tensor, respectively. The structure of the feature extraction module is illustrated in Fig.\ref{superpoint}. 
Features with score tensors and descriptors are filtered using a threshold \( T_f \).
However, a fixed confidence threshold can result in most extracted features being filtered out as the confidence of each feature point diminishes in challenging scenes. 

To address this issue, we introduce a self-adaptive strategy to dynamically adjust \( T_f \) according to the scene, enabling more robust feature extraction in challenging scenarios. The adaptive threshold mechanism considers two factors. The first is intra-feature relationship which refers to the distribution of feature points across all positions within a frame. We calculate the expectation \( E \) and variance \( \sigma^2 \) of the confidence distribution of the feature points within each frame. Assuming the confidence distribution follows a Gaussian distribution, the number of feature points within the range \( [E + \frac{\sqrt{\sigma^2}}{2}, +\infty) \) should remain constant, ensuring consistent feature extraction and avoiding situations where fixed thresholds may fail to extract sufficient feature points. The second is the inter-frame feature relationship which assesses the match quality between the current frame and adjacent frames. Since the interval between each frame is small, the matching quality of the previous frame can roughly reflect the current frame's matching quality. Let \( m \) denote the number of matched features between the current frame and adjacent frames. A decrease in matches signals a weak texture area or significant motion, indicating the need for extracting more feature points for subsequent tracking. Conversely, a strong match suggests a rich texture area, allowing for reduced feature extraction and lower computational costs. The adaptive threshold calculation formula is given in eq.\ref{eq:2}, where \(\mu_1\) and \(\mu_2\) are hyperparameters. Following the adaptive thresholding process, we index the positions \( (u_k, v_k), k \in m \) of feature points and obtain the descriptor tensor \( f_k, k \in m \). 
\begin{equation}
    T_f =E+\frac{\sqrt{\sigma ^2} }{2} +\mu_1\frac{1}{(1 + exp(-\mu_2m))}.
    \label{eq:2}
\end{equation}
\subsection{Feature Matching and Front-end}
\subsubsection{Feature Matching network}
Given two sets of feature points extracted from images $I_A$ and $I_B$, where $M$ and $N$ feature points are extracted, respectively, we denote the features of image $I_A$ as $\mathcal{A} = \{ f_i^A \}, i \in M$ and those of image $I_B$ as $\mathcal{B} = \{ f_j^B \}, j \in N$. Each feature point descriptor is normalized to fall within the range of 0 to 1. After normalization, the feature points are fed into the learning-based feature matching network, which outputs the soft partial assignment matrix $\mathbf{P} \in [0, 1]^{M \times N}$. The element $p_{ij} \in \mathbf{P}$ represents the likelihood that the $i$th feature point of $I_A$ matches the $j$th feature point of $I_B$. In our implementation, LightGlue is utilized for its capability to dynamically adjust the network depth and terminate early based on the input image's complexity. This feature significantly reduce the running time and memory consumption, making it highly suitable for SLAM tasks that require real-time performance.

\subsubsection{Tracking combined with learning-based matching}
Feature matching between adjacent frames relies on feature descriptor similarity of the query and target images. To speed up and reduce costs, the search area in the target image is restricted based on estimated positions predicted by camera motion. With short intervals between frames, VO assumes uniform motion, while VIO (Visual Inertial Odometry) uses IMU integration. Obviously, rapid acceleration or rotation can reduce this strategy's reliability due to pose prediction deviations. To address this issue, learning-based method offers a viable solution. By directly comparing features between current and previous frames in real-time, the matching network enhances robustness to aggressive motion, ensuring more reliable feature matching despite sudden changes in camera dynamics. In our implementation, LightGlue, equipped with a GNN (graph neural network) submodule capable of cross-view matching, is employed.

\begin{figure}[tb]
  \centering
  \includegraphics[width=0.95\linewidth]{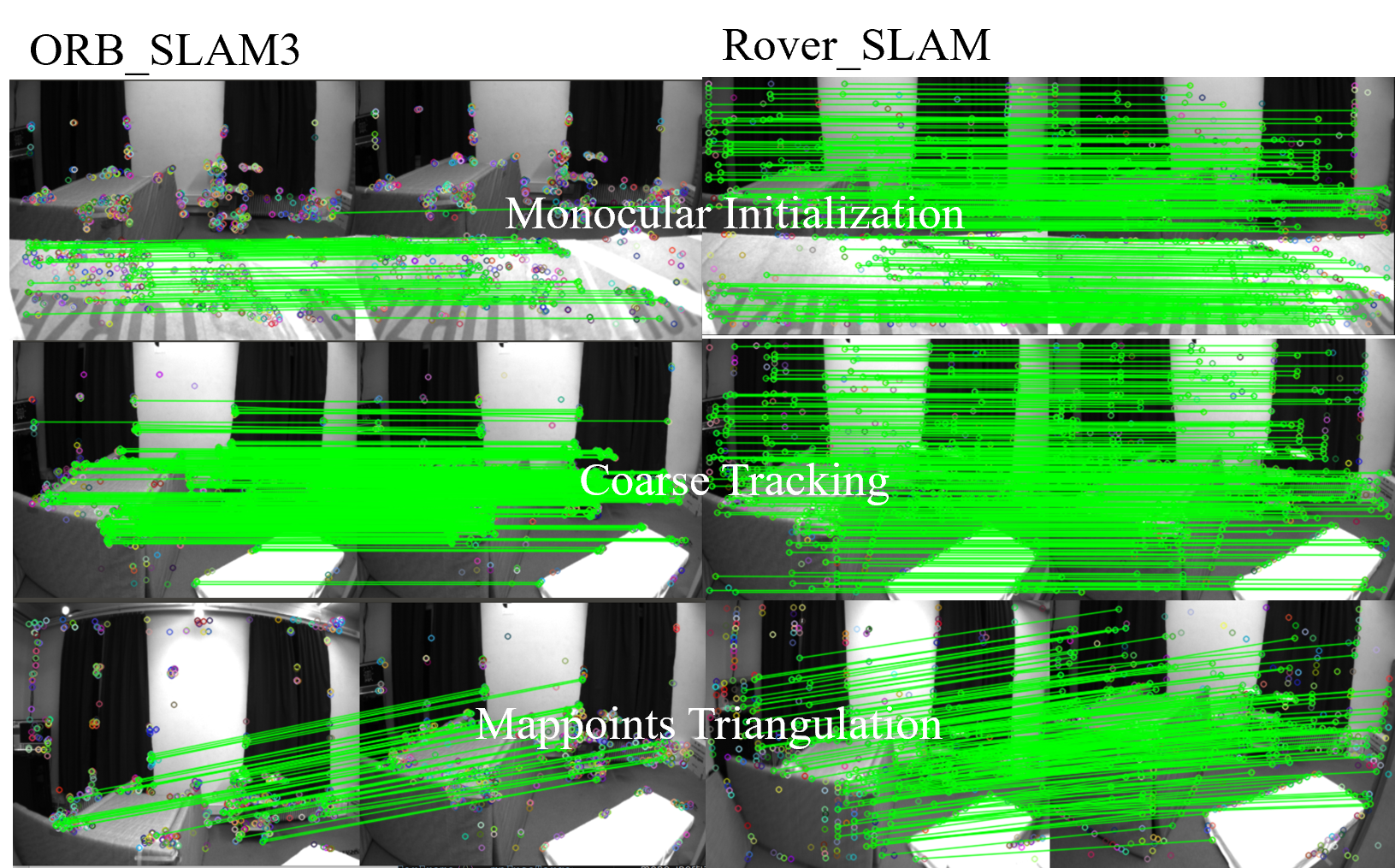}
  \caption{Comparison of matching performance in different stages. The first and second columns show the matching results of ORB-SLAM3 and Rover-SLAM, respectively. In the baseline system, coarse tracking utilizes project-based matching method, while monocular initialization and map points triangulation utilize DBoWs-based matching method.}
  \label{matchfigure}
\end{figure}

If continuous tracking fails, feature matching resorts to the latest keyframe, requiring descriptor-based matching. Consequently, feature descriptor-based matching becomes necessary. For example, in ORB-SLAM and VINS-Mono, the BoW (Bag-of-Word) algorithm expedites matching between current and reference keyframes for re-localization. However, the BoW approach converts spatial information into statistical information based on visual vocabulary, potentially losing accurate spatial relationships between feature points. To overcome these challenges, a learning-based matching approach is much more suitable. This transition has significantly increased the likelihood of successful tracking under large-scale transformations, thereby improving the accuracy and robustness of the tracking process. The effectiveness of different matching methods is illustrated in Fig. \ref{matchfigure}, where learning-based matching outperforms projection-based and BoW methods in ORB-SLAM3, with more evenly distributed and stable feature points.
\subsubsection{local mapping combined with learning-based method}
\begin{figure}[tb]
    \centering
    \includegraphics[width=1\linewidth]{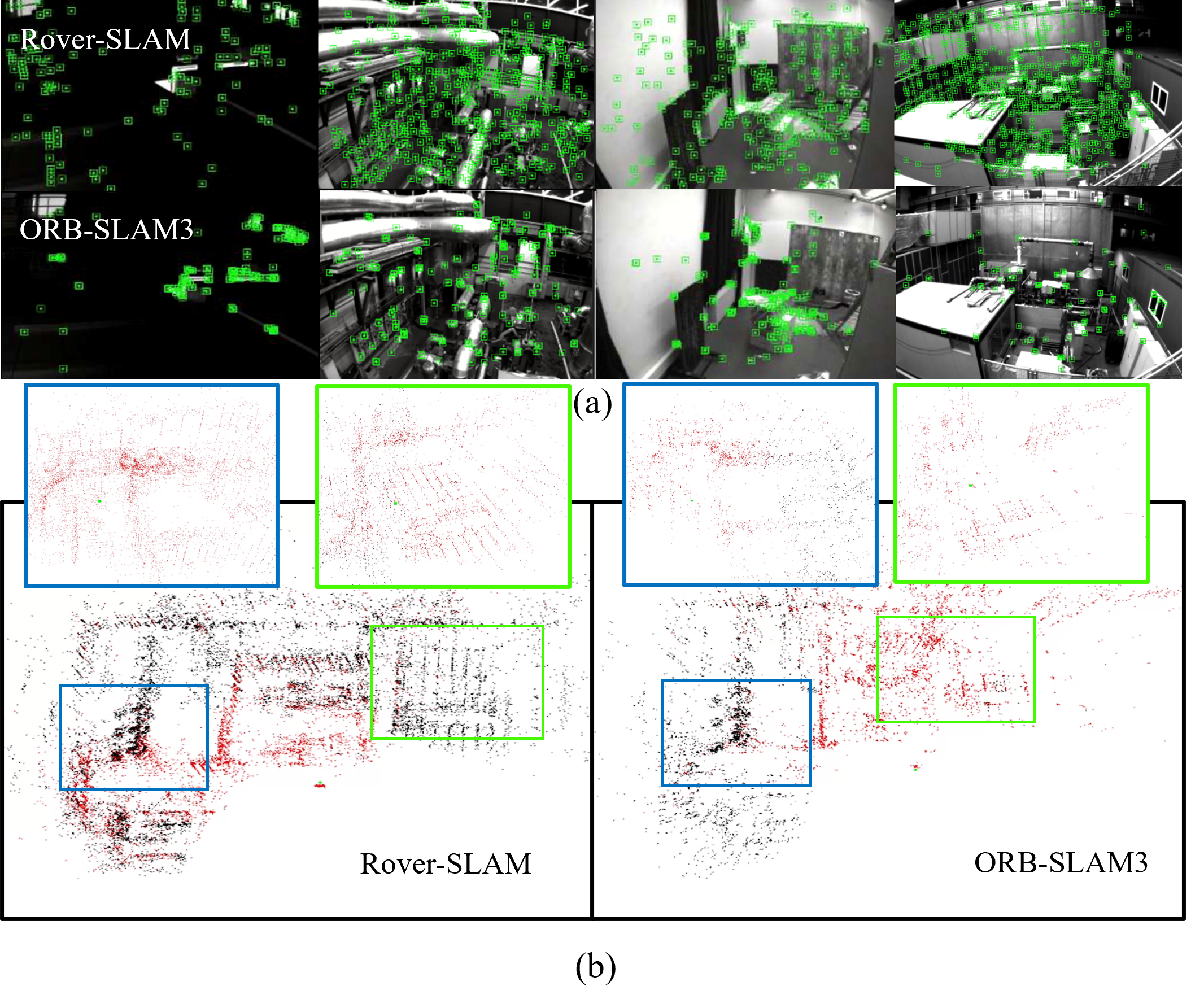}
    \caption{Map points comparison between Rover-SLAM and ORB-SLAM3. (a) shows a comparison of the reconstructed sparse point clouds, with rectangular of different colors highlighting the corresponding regions with significant differences. (b) shows a comparison of tracked map points.}
    \label{mappoint comparsion}
\end{figure}

In the local mapping thread, new map points are triangulated using the current keyframe and its neighboring keyframes. To obtain more accurate map points, it is essential to match with keyframes that exhibit larger parallax.
By employing learning-based matching and triangulating the matched points, more comprehensive and higher-quality map points can be obtained. This enhances the local mapping capability by establishing additional connections between keyframes and constructing a more robust essential graph. This, in turn, facilitates better joint optimization of the poses of co-visible keyframes and map points. The impact of triangulating map points is depicted in Fig. \ref{mappoint comparsion}. It can be observed that compared to ORB-SLAM3, the map points constructed by Rover-SLAM exhibit superior structural information of the scene. Additionally, they are distributed more evenly and extensively.
\subsection{Loop Closing}
\subsubsection{Deep bag-of-word descriptor}
The BoW approach used in loop closure is a visual vocabulary-based method, adapted from the bag-of-words concept in natural language processing. It starts with offline dictionary training, clustering feature descriptors from the training image set into \( k \) sets using K-means, forming the first level of the dictionary tree. This process is recursively applied within each set, resulting in a final dictionary tree with depth \( L \) and branch number \( k \). Each leaf node in this visual vocabulary tree is called a word.

During algorithm execution, BoW vectors and feature vectors are generated online from all features of the query image. Mainstream SLAM frameworks prefer manually set binary descriptors for their small memory requirements. To further enhance efficiency, we use a Gaussian distribution with an expected value of $0$ and a standard deviation of $0.07$ for the SuperPoint descriptor in our system. Consequently, the $256$-dimensional floating-point SuperPoint descriptor $d$ can be binary encoded to improve the indexing speed of visual place recognition. The binary encoding formula is given in e.q.\ref{eq:4}, where $d_i$ is the $i$th element of descriptor $d$, $0 \leq i < 256$.
\begin{equation}
d_i=\begin{cases}
 1, & \text{ if } d_i\ge 0\\
 0,& \text{ otherwise}.
\end{cases}
\label{eq:4}
\end{equation}
\subsubsection{Loop correction}
Loop closure in SLAM typically involves three key stages: seeking initial loop closure candidates, verifying loop closure keyframe, and performing correction by global bundle adjustment (BA).
First, initial loop closure candidates are identified using the previously trained BoW model. Keyframes that share common words with the current frame \( K_A \), but are not co-visible with it, are identified. The collective score of the co-visible keyframes associated with these candidate keyframes is computed. From the top \( N \) groups with the highest scores, the keyframe with the highest score is selected, denoted as \( K_B \).

Next, the relative pose transformation \( \mathbf{T}_{AB} \) from \( K_B \) to the current keyframe \( K_A \) must be determined. Similar to ORB-SLAM3, we uses BoW-based feature matching to match the current keyframe with the candidate keyframe \( K_B \) and its co-visible keyframe \( K_{C} \). It is worth noting that the learning-based matching algorithm significantly enhances matching efficiency, yielding high-quality map point correspondences. Finally, the Sim(3)(3D similarity transformation) transformation is solved to ascertain the initial relative pose \( \mathbf{T}_{AB} \). 

\section{Experiment}
\subsection{Implementation Details}
In this section, we evaluate the performance of Rover-SLAM with various sensor configurations through both qualitative and quantitative analyses. The assessments cover aspects such as pose estimation accuracy, running time, and robustness. To comprehensively evaluate Rover-SLAM's robustness, we tested our proposed method on three datasets: EuRoc\cite{euroc}, TUM-VI\cite{tumvi}, and our own dataset collected from real-world environments with various challenging conditions. All experiments were conducted on a computer equipped with an Intel Core i7-11700 CPU, 32GB of RAM, an NVIDIA RTX 3080 GPU, and running Ubuntu 18.04. To ensure flexibility, scalability, and real-time performance of Rover-SLAM, we deployed the deep learning models using the ONNX Runtime framework. We directly take the pre-trained SuperPoint and LightGlue models without any fine-tuning training. 
\subsection{Accuracy Comparison}
 For the pose estimation accuracy evaluation of Rover-SLAM, the open source EuRoc dataset is utilized. The EuRoc dataset encompasses recording scenes from office and factory environments and is categorized into three difficulty levels: easy (MH01/2, V101, V201), medium (MH03/4, V102, V202), and difficult (MH05, V103, V203). The difficult level is characterized by variable lighting conditions, severe shaking, and dark environments, offering a range of challenging scenarios. Since the EuRoc dataset provides both binocular images and inertial data, it allows us to test different sensor configurations, including monocular-inertial, stereo-inertial, and monocular modes. Notably, conducted extra in-depth tests on the monocular-inertial mode using TUM-VI dataset, as it is the most commonly used sensor type in robotic applications. Additionally, we selected the open-source versatile vSLAM solution, ORB-SLAM3, for comprehensive comparison.
\begin{figure}[tb]
    \centering
    \includegraphics[width=\linewidth]{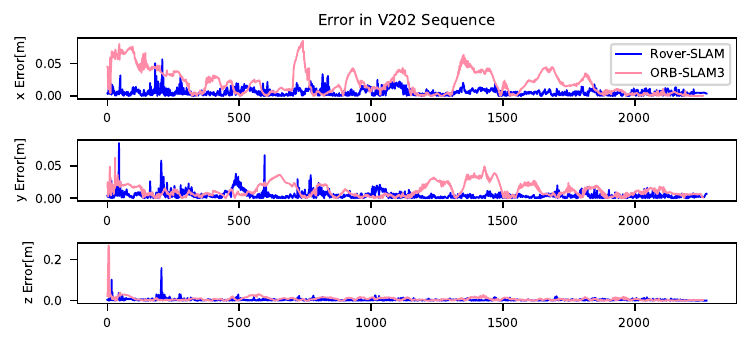} 
    \vspace{-1em} 
    \includegraphics[width=\linewidth]{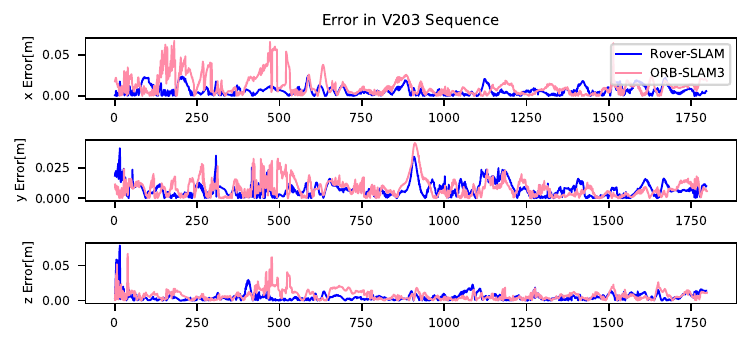} 
    \caption{Comparison of tracking error in EuRoc's V202 sequences between Rover-SLAM and ORB-SLAM3}
    \label{viserror}
\end{figure}
\subsubsection{Monocular-inertial tests}
In the \textbf{EuRoc dataset tests}, we selected several SOTA SLAM methods as contrast objects, including traditional visual-inertial SLAM methods like VI-DSO\cite{dsovi}, VI-ORB\cite{viorb}, OKVIS\cite{okvis}, VINS-Mono\cite{vins} , ORB-SLAM3\cite{orbslam3}, the hybrid SLAM algorithm SP-Loop\cite{Wang_Xu_Fan_Xiang_2023}, and the SOTA end-to-end deep learning SLAM system DVI-SLAM\cite{peng2023dvi}. We compared the trajectories from both quantitative and qualitative perspectives. The quantitative results of the ATE(absolute trajectory errors) in terms of RMSE(Root Mean Squared Error) are presented in Table \ref{vinscompare}. Rover-SLAM exhibited significant advantages over other approaches, leading in 8 out of the 11 sequences, with an average superiority of over $0.08m$ compared to ORB-SLAM3. It also outperformed other traditional algorithms by a large margin. In comparison to other deep learning-based SLAM systems, Rover-SLAM exhibited notable improvements in state estimation accuracy. This can be attributed to the extraction and matching of more stable and distinctive feature points, resulting in more complete and accurate estimations. The visualizations of the trajectory and error in each direction are shown in Fig.\ref{viserror}. When ORB-SLAM3 encounters challenging areas, the tracking becomes unstable and the tracking error increases. In contrast, Rover-SLAM maintains more stable and accurate positioning in these challenging scenarios.
\begin{table*}[tbh]
  \caption{RMSE$[m]$ of ATE comparison with SOTA VINS methods on EuRoc dataset. $\times$ denotes failure of the corresponding approach for this sequence. The best result is highlighted in bold.}
  \centering
  \label{vinscompare}
  \begin{tabular}{cccccccccc}
    \toprule
    \textbf{Dataset} & VI-DSO & VI-ORB & OKVIS & VINS-Mono & ORB-SLAM3& DVI-SLAM & SP-Loop &IMPS & Ours\\
    \midrule
    V101 & 0.109 & 0.095 & 0.090 & 0.146 & 0.091 & 0.074 & 0.042 & 0.090 & \textbf{0.034} \\
    V102 & 0.067 & 0.063 & 0.122 & 0.311 & 0.063 & 0.114 &0.034 & 0.061 & \textbf{0.019} \\
    V103 & 0.096 &  $\times$ & 0.196 & 0.329 & 0.066 & 0.083& 0.082 & 0.061 & \textbf{0.031}\\
    V201 & 0.075 & 0.081 & 0.168 & 0.124 & 0.076 & 0.091 & 0.038 & 0.074 & \textbf{0.024} \\
    V202 & 0.062 & 0.080 & 0.182 & 0.277 & 0.058 & 0.045 & 0.054 & 0.056 & \textbf{0.014}\\
    V203 & 0.204 & 0.114 & 0.305 & 0.323 & 0.063 & 0.072 & 0.100 & 0.065 & \textbf{0.018}\\
    MH01 & 0.074 & 0.105 & 0.292 & 0.177 & 0.044 &0.063& 0.070 & \textbf{0.039} & 0.048\\
    MH02 & \textbf{0.044} & 0.067 & 0.361 & 0.183 &0.083& 0.066 & 0.044 & 0.058 & 0.049\\
    MH03 & 0.124 & 0.040 & 0.267 & 0.404 & 0.044 &0.101& 0.068 & 0.040 & \textbf{0.039}\\
    MH04 & 0.112 & 0.075 & 0.366 & 0.394 & 0.082 &0.187& 0.100 & 0.054 & \textbf{0.052}\\
    MH05 & 0.121 & 0.152 & 0.396 & 0.382 & 0.064 &0.163& 0.09 & 0.058 & \textbf{0.048}\\
    \bottomrule
  \end{tabular}
\end{table*}
\begin{table}[tb]
    \caption{RMSE$[m]$ of ATE with SOTA VINS methods on TUM-VI dataset.}
    \centering
    \begin{tabular}{c c c c c || c}
    \toprule
    \textbf{Dataset} & VINS-Mono & ORB-SLAM3 & IMPS & Ours &length\\
    \midrule
       corridor1 & 0.63 & 0.23 & 0.19 & \textbf{0.086} & 305\\
       corridor2 & 0.95 & \textbf{0.04} & 0.04 & 0.062 & 322\\
       corridor3 & 1.56 & 0.40 & 0.35 & \textbf{0.039} & 300\\
       corridor4 & 0.25 & 0.16 & 0.15 & \textbf{0.007} & 114\\
       corridor5 & 0.77 & 0.31 & 0.31 & \textbf{0.009} & 270\\
       magistrale1 & 2.19 & 0.20 & 0.24 & \textbf{0.18} & 918\\
       magistrale2 & 3.11 & 0.46 & 0.27 & \textbf{0.23} & 561\\
    \toprule
    \end{tabular}
    \label{tumvicomapre}
\end{table}
\begin{figure}[bt]
    \centering
    \includegraphics[width=1\linewidth]{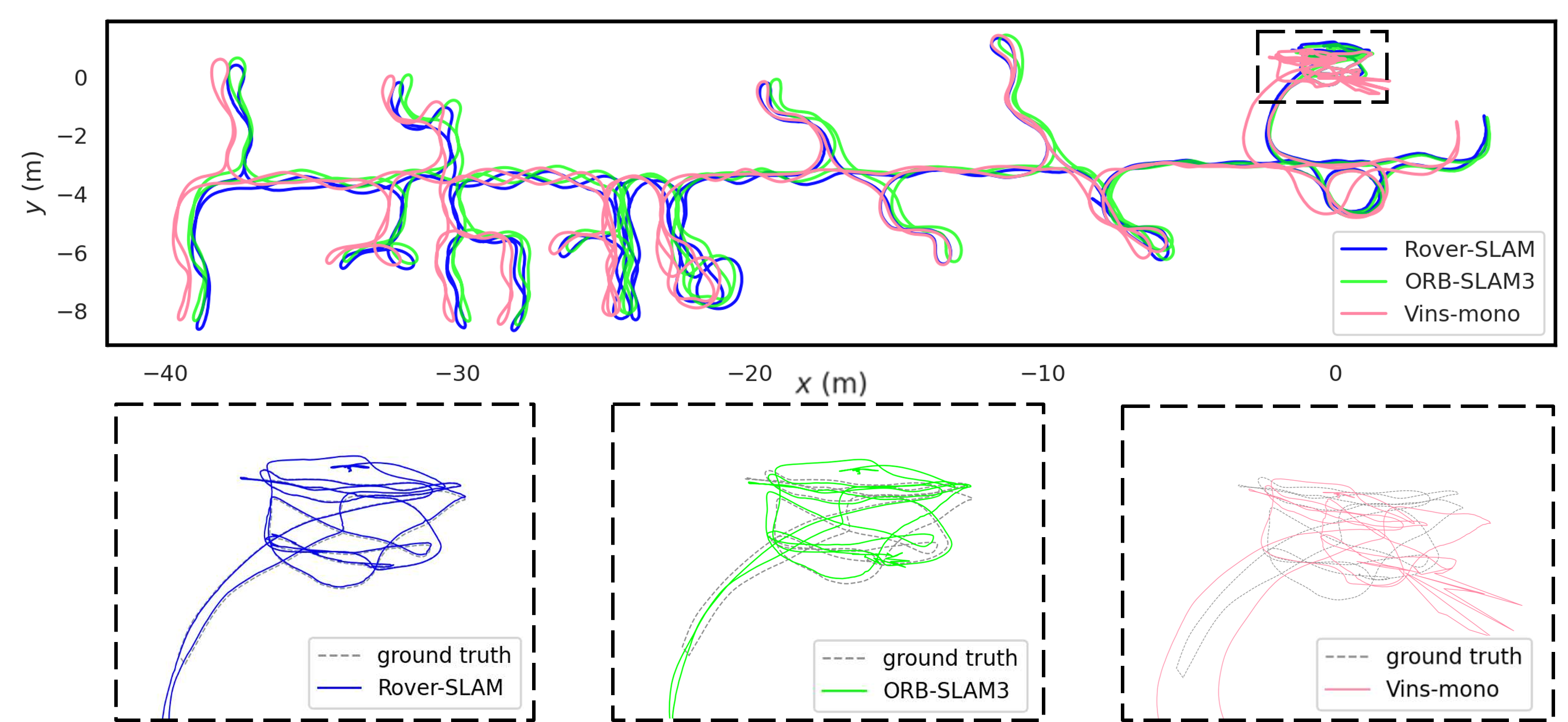}
    \caption{Representative trajectory visualization of Rover-SLAM on the TUM-VI dataset. In this test, only the last room contains the groundtruth (within the dashed box), and the details of the trajectory are shown at the bottom of the figure.}
    \label{fig:enter-label}
\end{figure}

In the \textbf{TUM-VI dataset tests}, we conducted experiments using ORB-SLAM3 to comprehensively illustrate the advantages of the proposed Rover-SLAM. This dataset was captured using handheld devices and evaluated with visual-inertial odometry. Groundtruth poses are provided only at the beginning and end of the trajectory, representing a small portion of the entire path. Therefore, we followed the methodology of ORB-SLAM3 to evaluate the accumulated drift over the entire trajectory. The results for monocular-inertial Rover-SLAM are presented in Table \ref{tumvicomapre}. The trajectory visualization on the corridor3 sequence is shown in Figure \ref{fig:enter-label}. Rover-SLAM achieved the lowest accumulated drift in most sequences and demonstrated superior performance. This result reflects the stability and reliability of our method.
\begin{table}[bt]
    \caption{RMSE$[m]$ of ATE comparison with SOTA stereo-vi methods on EuRoc.}
    \centering
    \begin{tabular}{c c c c c c}
    \toprule
    \textbf{Dataset} & VINS-Fusion & BASALT & Kimera & ORB-SLAM3 & Ours\\
    \midrule
       MH01 & 0.166 & 0.080 & 0.080 & 0.036 & \textbf{0.024}\\
       MH02 & 0.152 & 0.060 & 0.090 & 0.033 & \textbf{0.023}\\
       MH03 & 0.125 & 0.050 & 0.110 & 0.035 & \textbf{0.019}\\
       MH04 & 0.280 & 0.100 & 0.150 & 0.051 & \textbf{0.047}\\
       MH05 & 0.284 & 0.080 & 0.240 & 0.082 & \textbf{0.051}\\
       V101 & 0.076 & 0.040 & 0.050 & 0.038 & \textbf{0.033}\\
       V102 & 0.069 & 0.020 & 0.110 & \textbf{0.014} & 0.018\\
       V103 & 0.114 & 0.030 & 0.120 & 0.024 & \textbf{0.022}\\
       V201 & 0.066 & \textbf{0.030} & 0.070 & 0.032 & 0.046\\
       V202 & 0.091 & 0.020 & 0.100 & 0.014 & \textbf{0.012}\\
       V203 & 0.096 & $\times$ & 0.190 & \textbf{0.024} & 0.032\\
       
    \toprule
    \end{tabular}
    \label{stereo-vi compare}
\end{table}
\subsubsection{Stereo-inertial tests}
For the stereo-inertial sensor setups, several representative stereo-inertial SLAM systems, including VINS-Fusion\cite{vinsfusion}, BASALT\cite{BASALT}, Kimera\cite{kimera}, and ORB-SLAM3\cite{orbslam3}, are chosen as contrast targets. The test results for stereo-inertial Rover-SLAM are presented in Table \ref{stereo-vi compare}. It can be observed that Rover-SLAM achieved the best performance in 8 out of the 11 sequences. The superior performance of our method across multiple sensor configurations underscores its robustness and high scalability.

\subsubsection{Monocular tests}
For the monocular sensor mode, we selected direct vSLAM DSO\cite{dso}, semi-direct vSLAM SVO\cite{svo}, sparse feature SLAM ORB-SLAM3, and dense feature SLAM DSM\cite{dsm} for comparison, all of which are representative visual SLAM systems. The ATE of the estimated discrete poses is shown in Table \ref{monoculartable}. Among the 11 sequences in EuRoc, Rover-SLAM achieved SOTA results in 9 of them. In sequence V203, characterized by significant motion blur and photometric changes, ORB-SLAM3's monocular mode failed to track features. However, due to Rover-SLAM's robust feature tracking and loop closure detection ability, it quickly relocalized or merged multiple loops, despite occasional tracking failures, minimizing their impact. 
\begin{table}[bt]
    \caption{RMSE$[m]$ of ATE comparison with SOTA monocular methods on EuRoc.}
    \centering
    \begin{tabular}{c c c c c c}
    \toprule
    \textbf{Dataset} & DSO & SVO & DSM & ORB-SLAM3 & Ours\\
    \midrule
       MH01 & 0.046 & 0.100 & 0.039 & 0.016 & \textbf{0.007}\\
       MH02 & 0.046 & 0.120 & 0.036 & \textbf{0.027} & 0.032\\
       MH03 & 0.172 & 0.410 & 0.055 & 0.028 & \textbf{0.022}\\
       MH04 & 3.810 & 0.430 & 0.057 & 0.138 & \textbf{0.014}\\
       MH05 & 0.110 & 0.300 & 0.067 & 0.072 & \textbf{0.038}\\
       V101 & 0.089 & 0.070 & 0.095 & 0.033 & \textbf{0.021}\\
       V102 & 0.107 & 0.210 & 0.059 & 0.015 & \textbf{0.015}\\
       V103 & 0.903 & $\times$ & 0.076 & \textbf{0.033} & 0.037\\
       V201 & 0.044 & 0.110 & 0.056 & 0.023 & \textbf{0.021}\\
       V202 & 0.132 & 0.110 & 0.057 & 0.029 & \textbf{0.016}\\
       V203 & 1.152 & 1.080 & 0.784 & $\times$ & \textbf{0.031}\\
       
    \toprule
    \end{tabular}
    \label{monoculartable}
\end{table}

    
\begin{figure}[b]
    \centering
    \includegraphics[width=0.9\linewidth]{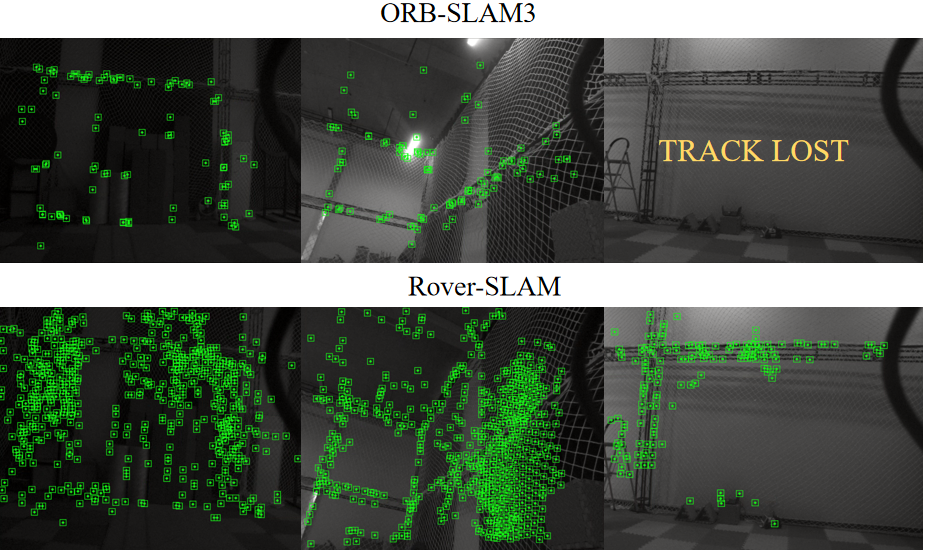}
    \caption{Tracking performance comparison of Rover-SLAM and ORB-SLAM3 on our dataset.}
    \label{zicai}
\end{figure}
\subsection{Tracking Robustness Evaluation}
\begin{figure} [tb]
    \centering
    \includegraphics[width=1\linewidth]{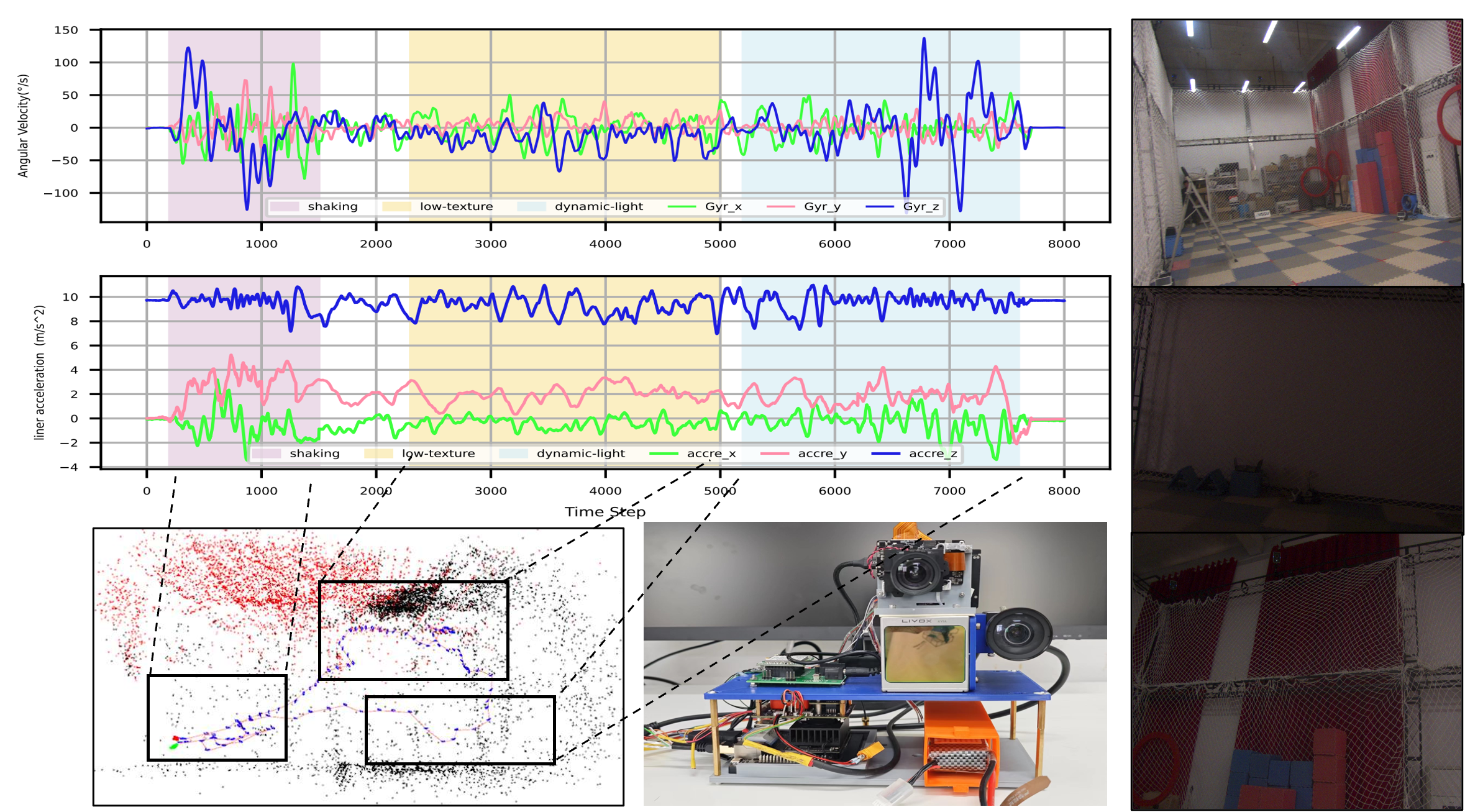}
    \caption{Self-collected dataset test. \textbf{Top-left: }Self-collected sequence characteristics in various challenge environments. The shaded area in different colors respectively represent different phases of shaking, low-texture, dynamic-light. \textbf{Bottom-left: }The estimated trajectory of Rover-SLAM on this sequence. \textbf{Bottom-middle: }
Diagram of data collection equipment. \textbf{Bottom-right: }Diagram of data collection environment. }
    \label{rosbag}
\end{figure}

To verify the robustness of a learning-based approach, we collected data using our own equipment in various challenging environments, including changes in illumination, low-texture, and shaking scenarios. Our handheld device for data sampling is shown in the bottom middle of Fig. \ref{rosbag}. It consists of a HIKVISON monocular camera with a resolution of $1440 \times 1080$, and an Xsense IMU with a frequency of $100 Hz$. 
The specifics of our sequence's details are shown in the top left of Fig. \ref{rosbag}, where we use different colors to indicate the challenging environments. In these conditions, as shown in Fig. \ref{zicai}, our method effectively extracts and tracks map points in low-texture areas and dark regions. The estimated trajectory, displayed in the bottom left of Fig.\ref{rosbag}, remains stable despite significant illumination changes. In contrast, ORB-SLAM3 loses track of map points and shows increasing drift or loses tracking altogether under such conditions. These results demonstrate that Rover-SLAM is robust enough to handle various challenging environments. 
\subsection{Time Costs Evaluation}
\begin{figure}[htb]
    \centering
    \includegraphics[width=1\linewidth]{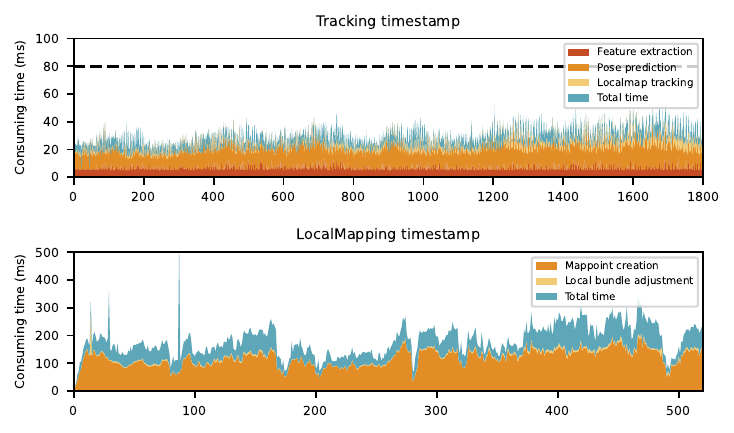}
    \caption{Time costs visualization of Rover-SLAM.}
    \label{runningtime}
\end{figure}

%
%
The SLAM system's tracking and local mapping threads consume most of the system resources, and the detailed runtime analysis of these two threads is shown in the Fig.\ref{runningtime}. Owing to the acceleration effect of deploying deep learning models with ONNX Runtime, it is evident that in the tracking thread, feature extraction for each frame takes approximately $7ms$, pose estimation takes around $10ms$, and the total tracking time is around $30ms$, allowing real-time tracking. Additionally, because the algorithm reconstructs a large number of map points, the map point creation module in the local mapping thread requires longer time, about $100ms$, and the total thread time is $180ms$. However, since this thread does not need to update every frame, its impact on the overall system's real-time performance remains limited.
\begin{table}[]
    \centering
    \caption{Ablation study. absolute trajectory error and relative pose error on EuRoc V201 sequence}
    \begin{tabular}{c c c c c}
    \toprule
     method  & ATE(RMSE)  & ATE(std) & RPE(RMSE) & RPE(std) \\
     \midrule
    ours & 0.0249 & 0.006 & 0.0047 & 0.0031\\
    MT & 0.0357 & 0.0156 & 0.0113 & 0.0102 \\
    LM & 0.0391 & 0.0247 & 0.0195 & 0.0133\\
    BL & 0.0760 & 0.0419 & 0.0262 & 0.0359\\
    \toprule
    \end{tabular}
    
    \label{ablationstudy}
\end{table}


\subsection{Ablation Study}

A series of tests was conducted to evaluate the efficacy of the modules proposed in our study. We utilized vSLAM system without integrated deep learning as a control and replaced the feature points of the corresponding module with ORB features, referred to as BL. Additionally, we replaced the tracking module of rover-SLAM with the original module. This setup was denoted as MT. Subsequently, the original triangulation-based mapping algorithm within the local mapping module, labeled as LM. The results of these experiments, conducted on the EuRoc dataset's V201 sequence, are summarized in the accompanying table \ref{ablationstudy}. In particular, we observed varying degrees of increase in ATE and RTE errors across the MT, LM, and baseline algorithms compared to our model. These findings underscore the importance of the proposed modules in enhancing the system's overall performance.

\section{Conclusion}


This letter introduces Rover-SLAM, a real-time, robust, and versatile visual-SLAM framework that integrates SOTA learning networks to improve the performance of traditional vSLAM solutions in challenging scenarios. In Rover-SLAM, the same features and matching networks are utilized for all tasks, i.e. tracking, mapping, relocalization, and loop closing. Additionally, novel schemes such as adaptive feature filtering, learning-based local map tracking, deep feature BoW, and versatile usage expansion are implemented to enhance efficiency, simplicity, reliability, and user-friendliness. Both quantitative and qualitative analyses confirm that Rover-SLAM signifies a significant advancement in the visual-based SLAM domain, delivering robust performance in complex environments while maintaining efficiency and scalability.

Revisiting the question "How to use deep-learning network in SLAM", we believe the answer depends on the application, but quite often the answer is \textit{hybrid}. The following reasons are specified:
\begin{itemize}
\item \textit{Real-time performance}: Deep learning networks designed for specific tasks rather than the complete SLAM task often do not require overly complex model structures, allowing for lightweight implementations. Leveraging GPUs ensures real-time performance of SLAM systems. It can be observed in the experiment that the feature extraction module in Rover-SLAM only requires $6ms$ to process a new coming frame.

\item \textit{Interpretability}: SLAM is a complex system that requires coordinated cooperation among multiple modules. Instead of integrating a large-scale model for the entire SLAM procedure, integrating various learning models across these modules separately helps in troubleshooting problems when the system malfunctions. In Rover-SLAM, each module's function can be tested and assessed independently, thanks to the distributed deep network implementation.

\item \textit{Generalization}: End-to-end SLAM methods need rich training samples and significant resource consumption to adjust network structures and retrain. However the generalization capability of these methods is still limited in different environments, complex scenarios and varying data input. Our approach, on the other hand, achieves satisfactory results with pre-trained models, greatly reducing resource consumption.

\item \textit{Theoretical foundation}: SLAM systems are founded on a solid base of probabilistic robotics theory, ensuring a thorough theoretical framework. Deep learning implementation should concentrate on specific issues rather of attempting to replace the entire SLAM system with deep learning models. By leveraging the strengths of both SLAM and deep learning, we can further improve the accuracy and robustness of SLAM systems in challenging real-world environments.
\end{itemize}

\bibliographystyle{IEEEtran}
\bibliography{IEEEabrv,ref}
\end{document}